\documentclass[letterpaper, 10 pt, conference]{ieeeconf}  

\IEEEoverridecommandlockouts                              

\usepackage{booktabs}
\usepackage{multirow}
\usepackage{graphicx}
\usepackage{algorithm}
\usepackage{algpseudocode}
\usepackage[hyphens]{url}
\usepackage{breakurl}

\title{\LARGE \bf
Decentralized LLM-Driven Coordination of Acoustic Robots for Contactless Object Manipulation
}

\author{Yingying Wang, Narsimlu Kemsaram, and Sriram Subramanian
\thanks{
Yingying Wang ({\tt\small yingying.wang.24@ucl.ac.uk}) and Sriram Subramanian ({\tt\small s.subramanian@ucl.ac.uk}) are with the Department of Computer Science, University College London, United Kingdom.
}%
\thanks{
Narsimlu Kemsaram ({\tt\small narsimlu.kemsaram@um.edu.my}) is with the Department of Artificial Intelligence, University of Malaya, Kuala Lumpur, Malaysia.
}%
}

\begin{document}
\sloppy

\maketitle

\begin{abstract}

Natural language interfaces can significantly reduce the complexity of interacting with multi-robot systems, especially in automation settings where non-expert users must issue high-level commands. In parallel, acoustic manipulation using ultrasonic phased arrays enables contactless object handling for contamination-sensitive applications such as healthcare, laboratory automation, and precision transport. However, the integration of large language models (LLMs) with distributed acoustic mobile robots remains largely unexplored. This paper presents a decentralized framework for natural language-driven coordination of acoustic robots for contactless object manipulation.
The proposed framework converts spoken instructions into executable multi-robot task plans through a pipeline that combines Whisper-based speech recognition, LLM-driven semantic parsing, structured JSON task representation, and distributed scheduling. The JSON schema encodes robot assignments, temporal dependencies, spatial constraints, and synchronization requirements for sequential, parallel, and synchronized execution. To support reliable coordination, we introduce a distributed synchronization protocol within a ROS2-based multi-agent architecture. The system is implemented on two TurtleBot3-based acoustic robots, each equipped with an ultrasonic phased array for contactless object transport.
The framework was evaluated in three task scenarios: sequential execution, parallel multi-robot transport, and synchronized cooperative manipulation. The system achieved task success rates of 96\% for sequential tasks, 86\% for parallel execution, and 70\% for synchronized collaborative transport. These results show that natural language commands can be consistently transformed into distributed robot actions for contactless manipulation.
The results demonstrate the feasibility of combining LLM-based task interpretation with decentralized multi-robot acoustic manipulation. While sequential and parallel tasks showed reliable performance, synchronized collaboration remained sensitive to alignment and timing errors. Overall, the proposed system highlights the potential of LLM-driven automation to improve human–robot interaction and enable flexible coordination in distributed robotic systems.

\end{abstract}

\section{INTRODUCTION}


Multi-robot systems (MRS) are increasingly deployed in automation scenarios requiring coordinated task execution across distributed robotic agents \cite{rizk2019cooperative}, \cite{ota2006multi}. Applications such as healthcare logistics, contamination-sensitive material handling, and micro-assembly demand flexible coordination strategies that allow humans to interact with robotic systems without specialized programming \cite{yan2013survey}, \cite{gautam2012review}. Traditional robotic control pipelines rely heavily on predefined command scripts or centralized task planners, which limit system adaptability and increase operational complexity \cite{8814594}, \cite{cortes2017coordinated}.

\begin{figure}[!htbp]
    \centering
    \includegraphics[width=0.45\textwidth]{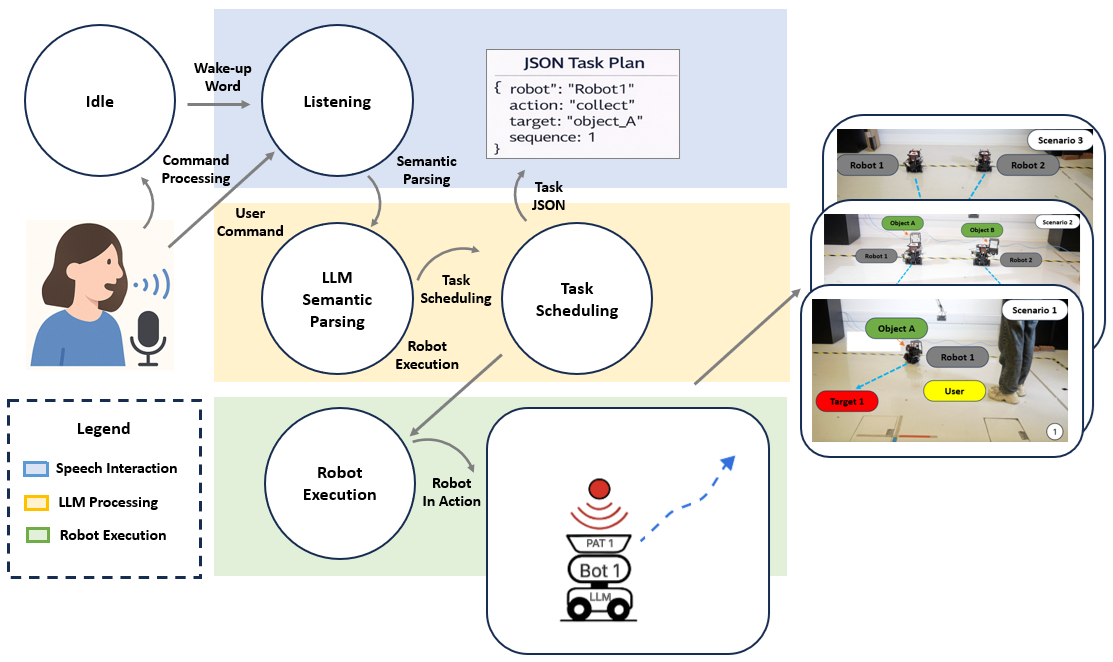}
    \caption{
    Proposed decentralized LLM-driven coordination of acoustic robots for contactless object manipulation. Finite-state control loop for natural language–driven multi-robot coordination. The system transitions from idle to listening upon wake-word activation. User commands are processed through an LLM-based semantic parsing module, which generates structured JSON task representations. The distributed scheduler assigns tasks to robots and executes them within the acoustic manipulation framework. Execution outcomes are monitored through real-time feedback, enabling adaptive task re-planning and continuous interaction.
    }
    \label{fig:fsm}
\end{figure}

Recent advances in LLMs create new opportunities for natural language–driven robot control \cite{wang2024lami}, \cite{qi2024advances}. By enabling robots to interpret instructions in everyday language, LLMs reduce the programming barrier for complex robotic systems \cite{zeng2023large}, \cite{wolfartsberger2025breaking}. However, most existing LLM-based robotic systems focus on single-robot command execution and lack mechanisms to coordinate multiple robots in distributed environments \cite{li2025large}, \cite{chen2024scalable}.


In parallel, acoustic manipulation technologies have emerged as a promising approach for contactless object handling \cite{stevens1899text}. Acoustic levitation systems generate radiation forces using ultrasonic phased arrays to trap and manipulate small objects without physical contact \cite{ochiai2014three}. Such systems are particularly valuable in environments where contamination must be avoided, including biomedical laboratories, pharmaceutical manufacturing, and healthcare applications \cite{andrade2016contactless}. While recent research has demonstrated acoustic manipulation using stationary platform \cite{nakahara2020contact}, integrating this capability with mobile robots introduces new opportunities for distributed contactless manipulation across large environments.

Despite these advances, several challenges remain. First, coordinating multiple robots through natural language commands requires structured task representations that capture temporal dependencies, synchronization constraints, and spatial relationships \cite{choudhury2022dynamic}. Second, integrating LLM-based reasoning with real-time robotic control introduces latency and reliability challenges, especially on resource-constrained edge devices \cite{friha2024llm}. Third, achieving synchronized acoustic manipulation across multiple mobile platforms requires precise coordination mechanisms and robust communication protocols \cite{nanzer2021distributed}.


To address these challenges, this paper proposes a decentralized LLM-driven automation framework for contactless multi-robot manipulation using acoustic mobile robots. The system integrates speech recognition and language-based task planning to convert natural-language commands into executable actions for a collaborative robot. A distributed scheduling architecture enables sequential, parallel, and synchronized multi-robot operations while maintaining robustness in dynamic environments (see Figure \ref{fig:fsm}).


Natural language interfaces have the potential to significantly reduce the complexity of interacting with robotic systems, particularly in environments where users may not have programming expertise. The integration of LLMs with distributed robotic systems enables a new paradigm of intuitive human–robot collaboration, where high-level instructions can be automatically translated into coordinated robot behaviors. By combining LLM-driven task planning with acoustophoretic manipulation, the proposed acoustic robot framework enables contactless multi-robot interaction that is particularly suitable for healthcare, laboratory automation, and contamination-sensitive environments. This work contributes toward the development of intelligent robotic systems that can understand human intent, coordinate autonomously, and safely manipulate objects without physical contact.

Our main contributions are: i) A distributed LLM-driven multi-robot automation architecture that integrates speech recognition and LLM-based semantic parsing to enable natural language-driven control of multiple acoustic robots, ii) A structured task representation for multi-robot coordination using a JSON-based task schema that encodes execution order, synchronization constraints, and robot assignments, enabling complex collaborative task execution, iii) A handshake-based coordination mechanism enables synchronized operation of multiple robots performing collaborative contactless transport, iv) Experimental validation of natural language-driven contactless manipulation with two acoustic robots and evaluated across three scenarios demonstrating sequential, parallel, and synchronized robotic manipulation tasks.

The remainder of this paper is organized as follows. Section II reviews related work on LLM-based human–robot interaction, multi-agent coordination, and acoustic manipulation. Section III presents the proposed system architecture and task representation framework. Section IV describes the system implementation, covering modules for speech processing, language parsing, multi-robot task scheduling, and action control. Section V evaluates the system across multiple evaluation scenarios and results. Finally, Section VI concludes the paper.


\section{RELATED WORK} 
\label{RelatedWork}

This section reviews prior research relevant to the proposed system, focusing on three main areas: i) Large language models for human–robot interaction, ii) Coordination strategies in multi-agent robotic systems, and iii) Acoustic manipulation.

\subsection{Large Language Models for Human–Robot Interaction}

Human–robot interaction research has increasingly focused on enabling intuitive interfaces that allow users to communicate with robots through natural language. Early human–robot interaction systems relied on rule-based dialogue systems or template-based command structures, which limited the range of instructions that could be interpreted \cite{bastianelli2017structured}.
Recent advances in LLMs have significantly improved natural language understanding and reasoning capabilities, enabling robots to interpret more complex instructions. Several studies have investigated integrating LLMs with robotic systems to translate natural language commands into structured task plans or executable robot actions \cite{kim2024survey}.
For example, LLM-based frameworks have been used to generate task plans for robotic manipulation, translate natural language instructions into robot API calls, and support interactive dialogue between humans and robots. These approaches demonstrate the potential of LLMs to reduce the programming burden associated with robot control \cite{liu2024enhancing}.
Despite these advances, several challenges remain. Many existing systems rely on centralized cloud-based architectures, which introduce latency and reliability concerns in real-time robotic applications \cite{xie2025centralized}. Additionally, most LLM-driven robotic systems focus on single-robot task execution and do not address the complexities of coordinating multiple robots performing collaborative tasks \cite{chen2024scalable}.
Furthermore, ensuring safe and deterministic robot behavior remains difficult when using generative language models. Techniques such as prompt engineering, constrained output schemas, and structured task representations have been proposed to improve reliability and enable robust integration with robotic control systems.

\subsection{Coordination in Multi-Agent Robotic Systems}

Multi-agent systems (MAS) enable multiple robots to collaborate in performing complex tasks that may be difficult or inefficient for a single robot to complete. Traditional multi-robot coordination approaches often rely on centralized planning architectures that assign tasks to individual robots and manage coordination through a central controller \cite{gautam2012review}.
While centralized approaches simplify system design, they have limitations such as reduced scalability, communication bottlenecks, and vulnerability to single points of failure. As a result, decentralized coordination strategies have gained attention in multi-robot systems \cite{chen2021decentralized}.
Decentralized coordination methods rely on local decision-making, distributed communication protocols, and synchronization mechanisms to enable cooperative task execution. Techniques such as distributed task allocation, auction-based scheduling, and consensus-based coordination have been widely studied in multi-robot research.
However, integrating natural language interaction with decentralized multi-robot coordination remains a relatively unexplored area. Existing LLM-based robotic frameworks rarely address distributed coordination mechanisms or real-time synchronization across multiple robots.




\subsection{Acoustic Manipulation and Contactless Transport}

Contactless manipulation technologies have attracted significant attention because they handle objects without physical contact, which is important in contamination-sensitive environments such as biomedical laboratories, pharmaceutical manufacturing, and micro-assembly systems \cite{vandaele2005non}, \cite{al2022non}. Among non-contact manipulation approaches, acoustic levitation has emerged as a promising technique for trapping, manipulating, and transporting objects using ultrasonic radiation forces generated by phased array of transducers \cite{ochiai2014three}, \cite{foresti2013acoustophoretic}.
Previous research has demonstrated the feasibility of acoustic manipulation at small scales using programmable phased-array systems that dynamically shape acoustic pressure fields \cite{8094247}, \cite{melde2016holograms}. These systems enable particle levitation, sorting, and trajectory control in air or liquid environments. For example, ultrasonic phased arrays have been used to manipulate microparticles and droplets through dynamic acoustic field control \cite{foresti2013acoustophoretic}.
Recent work has explored scaling acoustic manipulation to larger objects and integrating it with robotic systems \cite{andrade2020acoustic}. Experimental demonstrations have shown that low-frequency ultrasonic transducer configurations can levitate centimeter-scale objects under controlled conditions. These results suggest the potential for mobile acoustic manipulation systems capable of operating in larger environments.
However, most existing acoustic levitation systems are stationary laboratory setups and require precise calibration and centralized control. Integrating acoustic manipulation with mobile robots remains relatively unexplored, especially in the context of distributed multi-robot coordination and intuitive human–robot interaction.





Although significant progress has been made in LLM-based human–robot interaction, acoustic manipulation, and multi-agent robotic coordination, these research areas have largely developed independently. Existing acoustic manipulation systems are typically limited to stationary platforms. 
This work addresses these challenges by proposing a decentralized multi-agent framework for contactless object manipulation using acoustic mobile robots.
\section{PROPOSED SYSTEM} \label{ProposedSystem}


This section presents a proposed decentralized, natural-language-driven framework for contactless object manipulation using acoustic robots. 

\subsection{System Concept}

The proposed system introduces a decentralized coordination framework for acoustic mobile robots that enables users to issue high-level natural language commands for contactless object manipulation. Unlike traditional robotic systems that rely on predefined scripts or centralized planners, the proposed framework transforms spoken instructions into executable multi-robot task sequences using large language models (LLMs). The overall objective is to reduce the need for domain-specific programming while enabling flexible multi-robot coordination in acoustophoretic robotic systems.

\subsection{System Architecture}

\begin{figure*}[!htbp]
    \centering
    \includegraphics[width=0.675\textwidth]{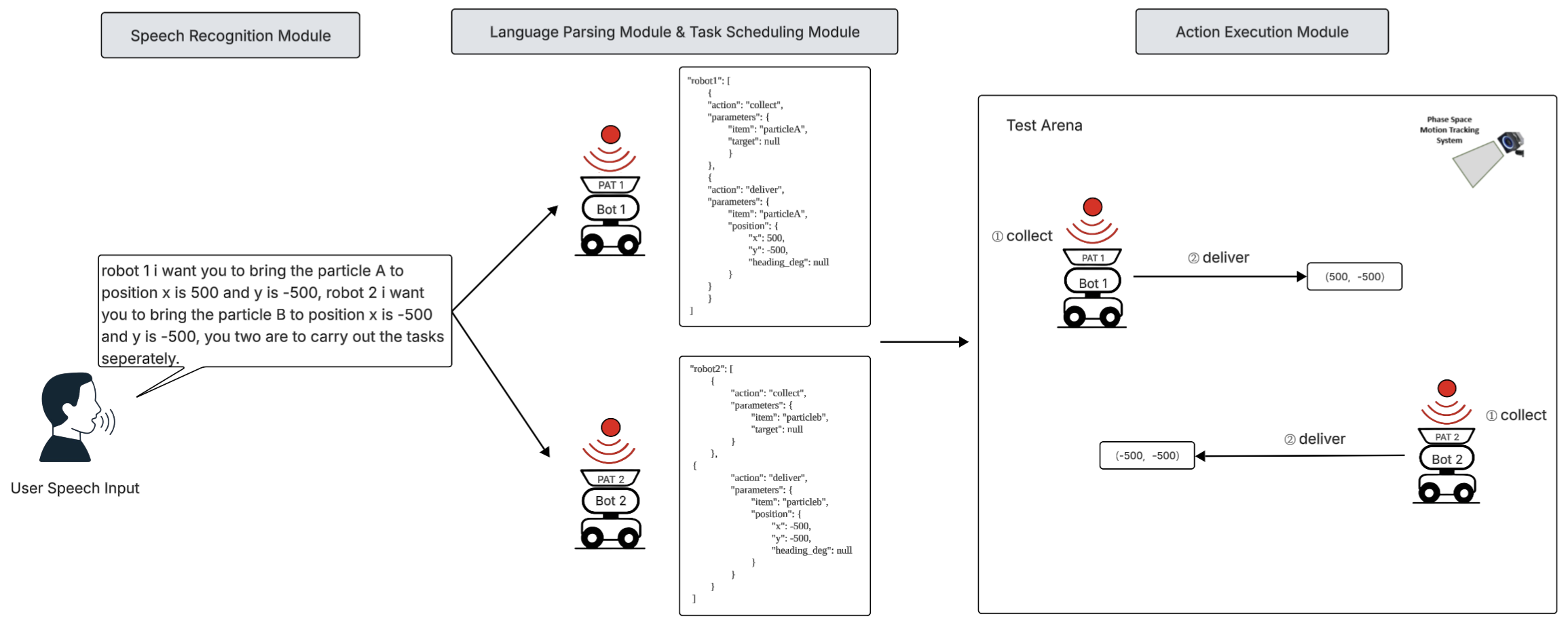}
    \caption{
    Architecture of the proposed decentralized natural language–driven multi-agent framework. Spoken user commands are first transcribed by the Whisper speech recognition module and then interpreted by a large language model to generate structured JSON task plans. A distributed scheduler assigns tasks to acoustic robots, which execute sequential, parallel, or synchronized actions within the test arena.
    }
    \label{fig:SystemArchitecture}
\end{figure*}

The overall architecture of the proposed framework is shown in Figure \ref{fig:SystemArchitecture}. The system comprises six functional modules that transform user input into coordinated robot actions: (i) speech recognition, (ii) LLM-based language parsing, (iii) structured task representation, (iv) distributed scheduling and coordination, (v) robot control and motion execution, and (vi) acoustic manipulation. Together, these modules form a closed-loop natural language-to-execution pipeline.




Natural language interaction begins with an onboard speech interface using the Whisper automatic speech recognition model. The speech module performs wake-word detection, voice activity detection, and speech-to-text transcription. Once the robot detects a predefined activation phrase, such as ``open robot system'', it switches from idle mode to command acquisition mode. The spoken command is then recorded and converted to text for semantic processing.
The language parsing module forms the core of the proposed system. It converts natural language instructions into structured machine-readable task representations using an LLM. To improve reliability, the parser uses a multi-stage prompting strategy with constrained action vocabularies and parameter validation. The parsed output is encoded as a structured JSON task plan containing robot assignments, action types, sequence dependencies, and synchronization constraints.

To support distributed execution, the framework introduces a structured task representation model and a distributed scheduler. Parsed task sequences are allocated across robots while preserving execution constraints. The scheduler supports sequential, parallel, and synchronous execution modes and coordinates robots through a three-way handshake protocol. This mechanism enables reliable synchronization without centralized control and supports dependency analysis, timeout detection, and fault recovery.
Low-level robot actions are executed through a ROS~2-based control framework. Each acoustic robot supports navigation primitives such as \texttt{move}, \texttt{rotate}, \texttt{navigate}, and \texttt{follow}. Robot localization is provided by the PhaseSpace motion capture system, which offers millimeter-level position estimates. Motion control combines coarse navigation, fine alignment, and heading correction, while an adaptive velocity controller reduces motion errors near target positions. To avoid collisions during multi-robot operation, robots broadcast motion intentions and dynamically allocate safe positions around shared targets.

A distinguishing capability of the proposed system is contactless object manipulation using acoustic levitation. Each robot is equipped with an $8\times8$ ultrasonic phased array that generates acoustic radiation forces for levitating and transporting small objects. The acoustic pressure field generated by the phased array can be approximated as

\begin{equation}
P(x,t)=\sum_{i=1}^{N} A_i \sin(\omega t + \phi_i)
\end{equation}

where $A_i$ is the transducer amplitude, $\omega$ is the acoustic frequency, $\phi_i$ is the phase offset, and $N$ is the number of transducers. By adjusting the phase offsets, the system creates stable acoustic potential wells that can trap and transport objects. In collaborative manipulation scenarios, two robots synchronize their acoustic fields to create a shared levitation region for distributed contactless transport.

\subsection{Control Flow and Execution Pipeline}

The proposed framework adopts a finite-state machine (FSM) to coordinate the full control pipeline from startup to task execution, as shown in Figure \ref{fig:fsm}. Upon wake-word detection, the system transitions from idle to listening mode, records the user command, and invokes the LLM to parse the instruction into structured task sequences. Parsed tasks are scheduled using sequence and synchronization logic and then dispatched for execution. During operation, the system provides real-time audio feedback. Deactivation or exit phrases return the system to idle or terminate the process.

\subsubsection{Idle State and Wake-Word Detection}

The system starts in an idle state and continuously monitors for predefined activation phrases through low-power audio processing. Wake-word detection enables hands-free command initiation and balances recognition accuracy with computational efficiency on resource-constrained embedded platforms.

\subsubsection{Speech Processing}

Upon wake-word activation, the speech subsystem uses voice activity detection to capture the spoken command and then transcribes it using Whisper. The resulting text is forwarded to the semantic parsing stage.

\subsubsection{Language Parsing}

The language processing module converts user commands into structured JSON task representations using a multi-stage parsing pipeline with error recovery. Initial parsing attempts use a low sampling temperature, while subsequent retries gradually increase temperature if parsing fails. This approach improves robustness against linguistic variability and ambiguity.

The parsing framework uses constrained prompts that define action vocabularies, parameter specifications, and coordination semantics. The parser supports navigation actions (\texttt{move}, \texttt{turn}, \texttt{navigate}), manipulation actions (\texttt{collect}, \texttt{deliver}, \texttt{transport}), and simple social behaviors (\texttt{speak}, \texttt{wait}). Spatial reference resolution handles expressions such as ``here'', ``there'', and relative positions through semantic mapping between user and robot coordinate frames.

Multi-robot coordination is encoded through structured dependency fields. The \texttt{sequence} field specifies ordered execution for sequential tasks, while the \texttt{sync\_group} field enables synchronous collaborative actions. Independent execution is the default unless explicit synchronization constraints are present.

\subsubsection{Distributed Scheduling and Coordination}

The scheduler bridges parsed tasks and execution and supports four coordination modes: single-robot sequential execution, multi-robot parallel execution, cross-robot ordered execution, and synchronous collaborative execution. Coordination is represented using \texttt{sequence} and \texttt{sync\_group} fields. A TCP-style three-way handshake ensures reliable task acknowledgment and synchronized start signals over ROS~2 wireless communication. Barrier synchronization logic guarantees simultaneous execution in synchronous mode, while dynamic timeouts and degradation protocols improve robustness under failures or disconnections.

\subsubsection{Robot Motion and Task Execution}

The robot controller dispatches low-level motion commands derived from the parsed tasks. Navigation combines open-loop motion control, heading calibration, and MoCap-based fine alignment. For conflict resolution, robots exchange intent information and allocate safe positions around shared targets using deterministic arrangements.
For synchronous collaborative transport, the \texttt{contactlessTransport} module enables coordinated dual-robot object manipulation. The execution pipeline includes formation planning, navigation calibration, and synchronized transport. In this mode, the robots first move into a back-to-back formation with fixed spacing, validate symmetry, exchange readiness signals, and then execute synchronized motion using a shared start timestamp. One robot acts as the leader to avoid conflicts, while the second robot follows the agreed timing signal. This process enables stable collaborative acoustic transport while supporting timeout handling and exception recovery.

\subsubsection{Feedback and State Management}

The system provides real-time audio feedback throughout execution to confirm task progress and completion status. Robust error handling ensures graceful degradation during failures and supports smooth transitions back to the listening state, enabling continuous human--robot interaction.

\section{EVALUATION and RESULTS}
\label{Evaluation}

This section evaluates the proposed natural language-driven multi-robot
framework on the acoustic robot platform across three key aspects:
1) execution performance under different task coordination modes,
2) acoustic levitation stability during robot operation, and
3) natural language understanding and speech recognition performance.
Experiments were conducted using two TurtleBot3-based acoustic robots
in a tracked indoor environment.



\subsection{Experimental Setup}

Experiments were performed using two TurtleBot3-based acoustic robots equipped with $8\times8$ ultrasonic phased-array transducer boards for contactless object manipulation. Each robot included an onboard microphone, speaker, and Raspberry Pi 4B embedded computing unit running Ubuntu 22.04 and ROS 2 Humble middleware. Speech commands were transcribed using Whisper, while robot localization was obtained through the PhaseSpace motion capture system, which provided millimeter-level position accuracy in the experimental arena. The robots
communicated through a ROS 2-based distributed communication network.

\begin{figure}[!htbp]
    \centering
    \includegraphics[width=0.40\textwidth]{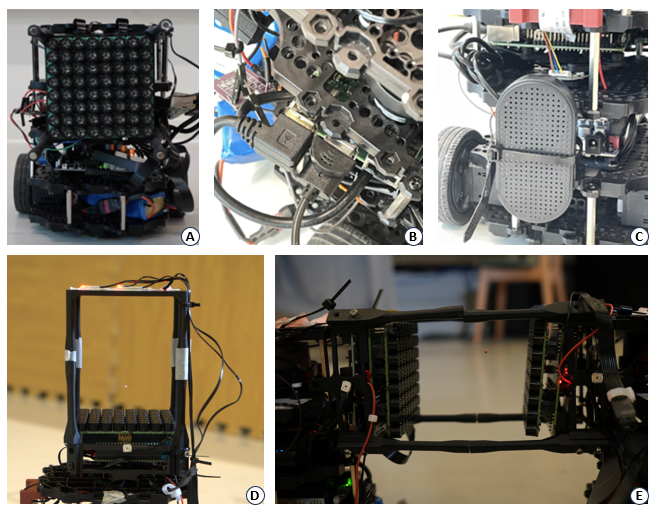}
    \caption{
    Experimental setup: Hardware components of the acoustic robot experimental platform used for contactless object manipulation, A) Front view of the acoustic mobile robot equipped with an ultrasonic phased-array module for acoustic radiation force generation, B) Rear view illustrating the onboard computing unit and wiring layout, including PhaseSpace LED marker configuration used for high-precision motion tracking in the experimental arena, C) Raspberry Pi 4B control unit integrated with USB microphone and speaker modules to enable natural language interaction, D) and E) Acoustics phased-array module responsible for generating acoustic pressure fields on single-sided phased array and face-to-face phased arrays to enable contactless object and cooperative contactless object transportation.
    }
    \label{fig:acoustobot_architecture}
\end{figure}

Figure \ref{fig:acoustobot_architecture} shows the hardware components of the acoustic robot experimental platform, including the mobile base, onboard control unit, motion-tracking markers, and ultrasonic phased-array module.


\subsection{Experimental Scenarios}



\begin{figure}[!htbp]
  \centering
  \includegraphics[width=0.40\textwidth]{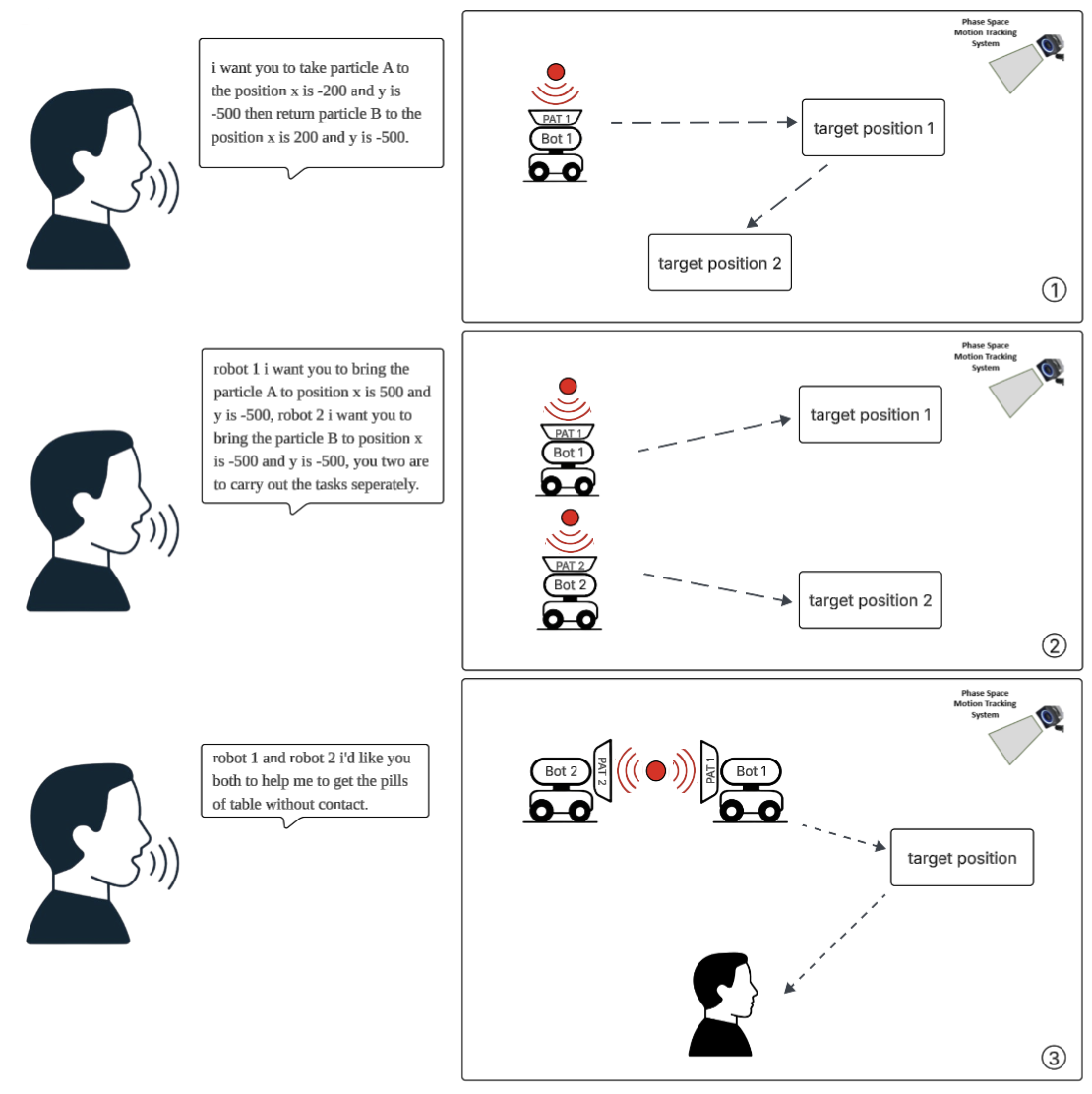}
    \caption{
    Natural language–driven multi-robot coordination scenarios for acoustic contactless manipulation. A user provides spoken instructions that are converted into structured task plans and executed by acoustic mobile robots in a shared test arena. The framework supports three coordination modes: (1) Sequential execution, where a single robot performs ordered transport tasks, (2) Parallel execution, where multiple robots independently transport objects simultaneously, and (3) Synchronous cooperative execution, where robots coordinate their motion and acoustic manipulation to jointly transport an object.
    }
  \label{fig:teaserDiagram}
\end{figure}

To evaluate system capabilities, three representative task coordination scenarios were considered (as illustrated in Figure \ref{fig:teaserDiagram}): 1) \emph{Sequential task execution}, where a single robot performs ordered object transport tasks based on natural language instructions, 2) \emph{Parallel task execution}, where two robots independently execute transport tasks simultaneously after receiving a shared instruction, and 3) \emph{Synchronous collaborative execution}, where two robots cooperatively transport a shared object using synchronized acoustic manipulation.
These scenarios represent increasing levels of coordination complexity, ranging from single-robot operation to cooperative multi-robot manipulation.

\subsubsection{Sequential Task Execution and Acoustic Levitation Stability Analysis (Scenario 1)}

\begin{figure}[!htbp]
    \centering
    \includegraphics[width=0.40\textwidth]{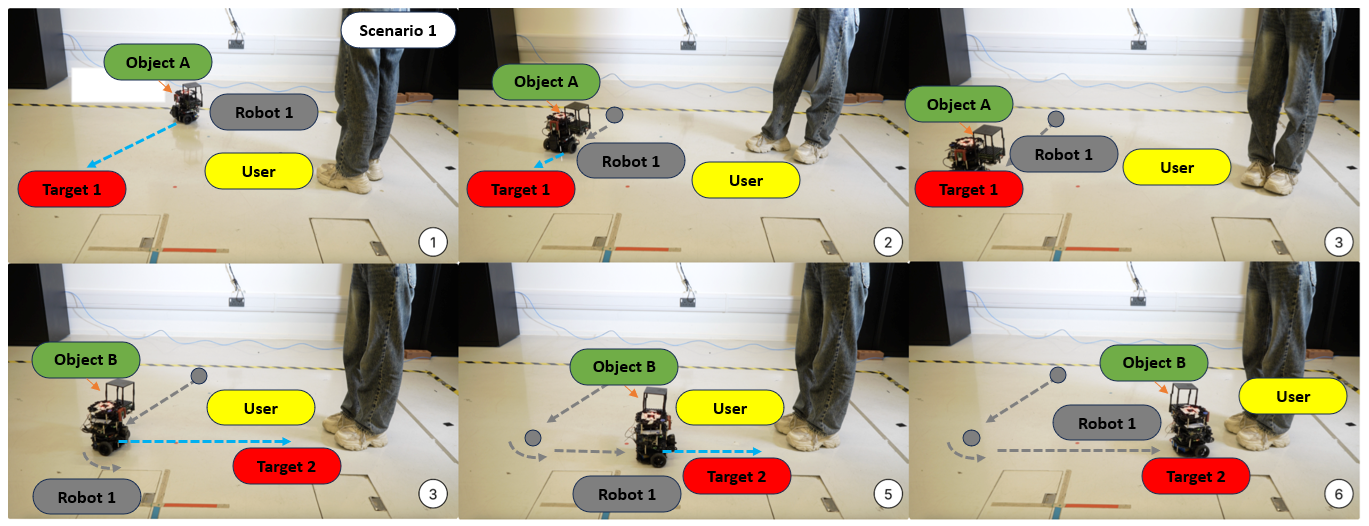}
    \caption{
    Sequential task execution scenario used for experimental evaluation (Scenario 1): A single acoustic robot executes two consecutive object transport tasks following natural language command interpretation. (1–3) The robot navigates toward the first target location and delivers Object A. (4–6) The robot then proceeds to the second target position and completes the transport of Object B.
    }
    \label{fig:seqTrajectory}
\end{figure}

\begin{figure}[!htbp]
    \centering
    \includegraphics[width=0.40\textwidth]{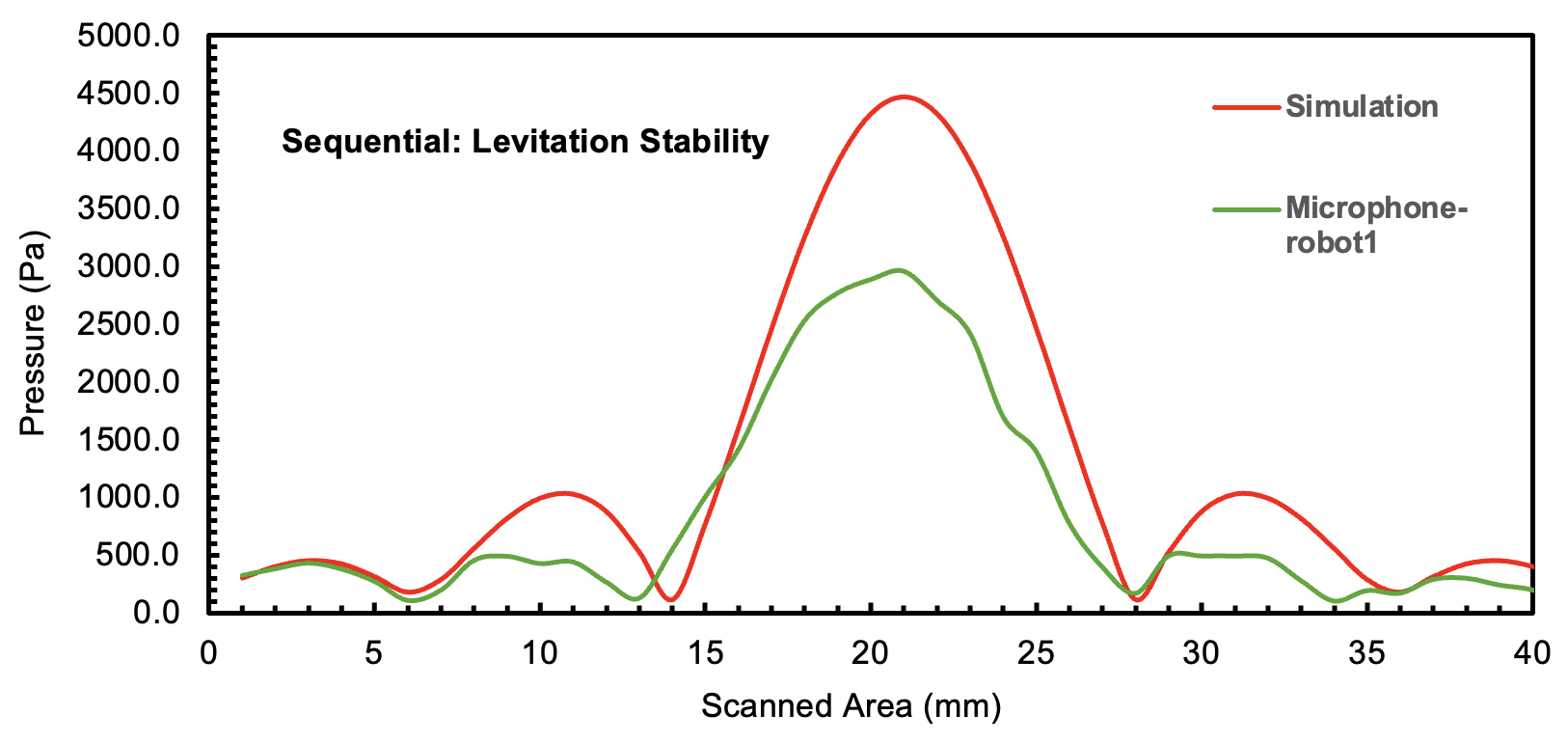}
    \caption{
    Sequential levitation stability analysis (Scenario 1): Comparison between MATLAB simulation (red curve) and experimental microphone measurements from Robot 1 (green curve) across the scanned acoustic field. The results show a similar pressure distribution pattern, validating the accuracy of the acoustic field simulation used for contactless object manipulation.
    }
    \label{fig:sequential_levitation}
\end{figure}

The sequential scenario evaluates whether a single acoustic robot can reliably execute ordered multi-step transport tasks derived from natural language input. As shown in Figure \ref{fig:seqTrajectory}, the robot receives a command describing two consecutive object transport tasks. The LLM-based parser converts the command into structured task representations, which are then executed sequentially by the scheduler.
During execution, the robot first navigates toward the location of Object A and transports it to the designated target position. After completing the first step, the robot proceeds to the second target and transports Object B. This experiment demonstrates that the framework can reliably interpret and execute ordered multi-step instructions using natural language commands.
To validate the acoustic manipulation process during sequential execution, simulated acoustic pressure fields were compared with experimental microphone measurements. Figure \ref{fig:sequential_levitation} shows that the MATLAB simulation closely matches the measured pressure profile from Robot 1 across the scanned region. The similarity between the simulated and measured acoustic field confirms the validity of the acoustic model used for contactless object manipulation in the single-robot scenario.



\subsubsection{Parallel Multi-Robot Execution and Acoustic Levitation Stability Analysis (Scenario 2)}

\begin{figure}[!htbp]
    \centering
    \includegraphics[width=0.40\textwidth]{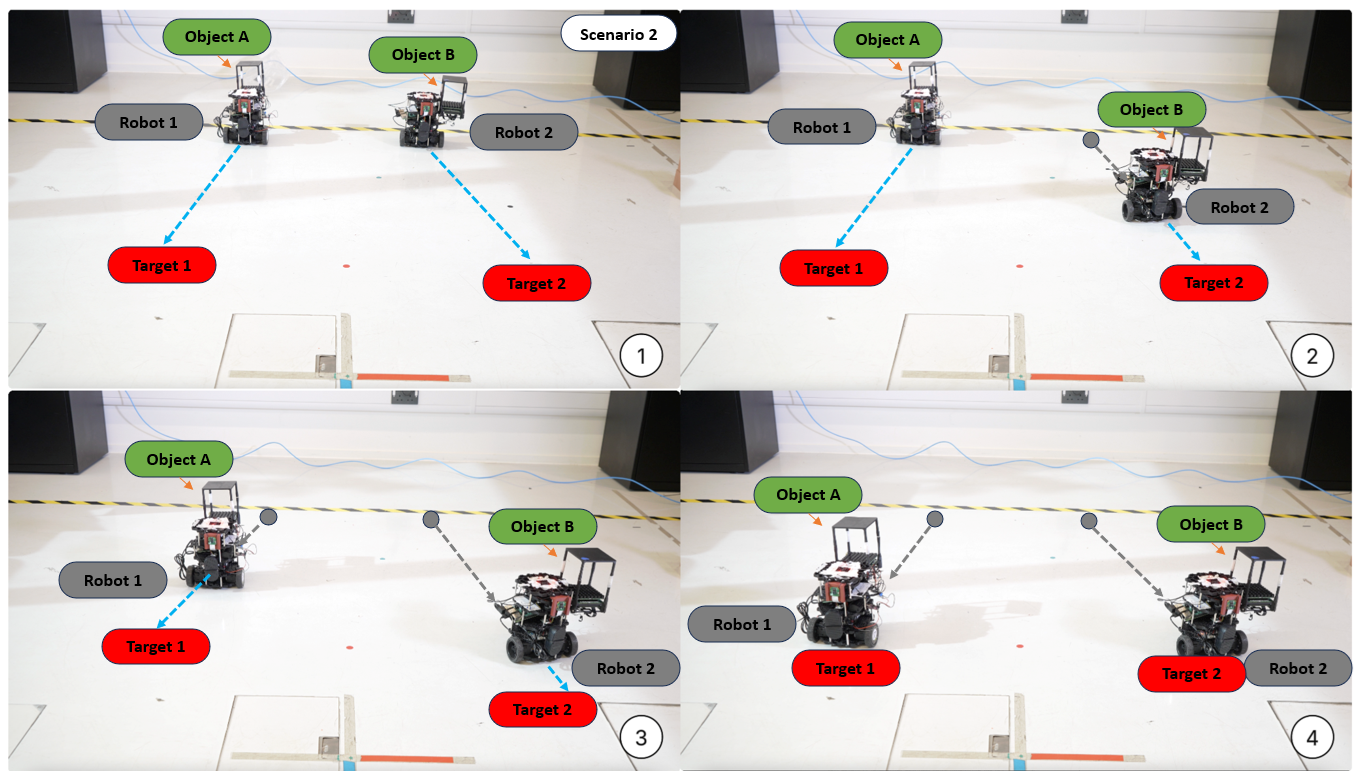}
    \caption{
    Parallel task execution scenario used for experimental evaluation of multi-robot coordination (Scenario 2): Two acoustic robots transport objects to different target positions simultaneously following natural language command parsing and distributed task scheduling. (1–2) Both robots navigate toward their assigned target locations. (3–4) Each robot independently completes its object transport task using acoustic manipulation.
    }
    \label{fig:paraTrajectory}
\end{figure}

\begin{figure}[!htbp]
    \centering
    \includegraphics[width=0.40\textwidth]{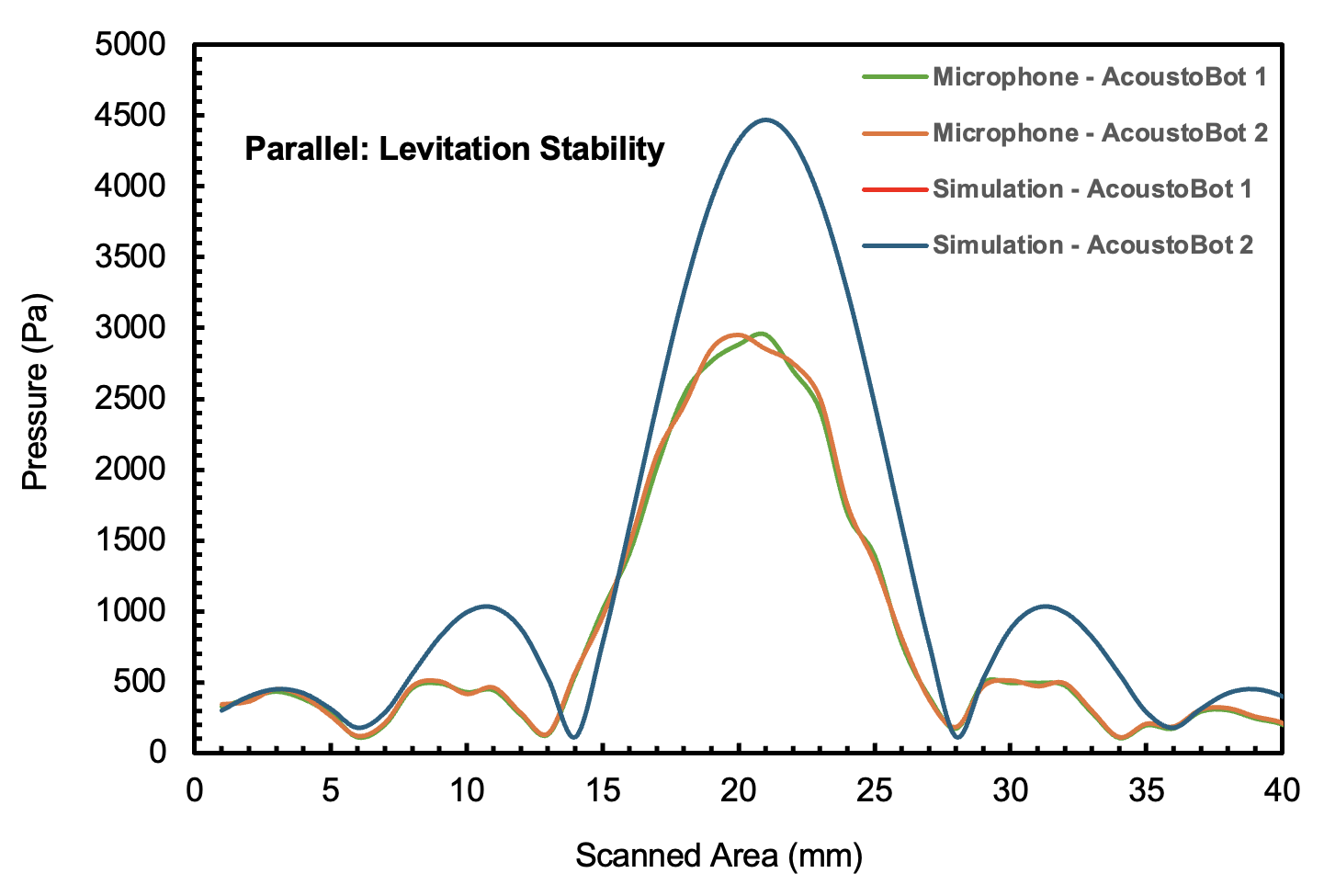}
    \caption{
    Parallel levitation stability analysis during multi-robot operation (Scenario 2): The figure compares acoustic pressure profiles obtained from MATLAB simulation and microphone measurements for two acoustic robots operating simultaneously. The results demonstrate consistent pressure distributions across the scanned area, validating the stability of acoustic levitation during parallel object transport tasks.
    }
    \label{fig:parallel_levitation}
\end{figure}

The parallel scenario evaluates decentralized task execution in which two robots operate simultaneously on separate object transport tasks. As shown in Figure \ref{fig:paraTrajectory}, the distributed task scheduler assigns individual tasks to each robot based on the parsed instruction, allowing both robots to execute independently in parallel.
Each robot navigates toward its assigned target location and completes the corresponding object transport task without waiting for the other robot. This experiment demonstrates that the proposed framework supports concurrent task execution and improves throughput under multi-robot operation.
Because simultaneous operation may introduce interference between acoustic fields, the pressure distributions generated by both robots were evaluated through simulation and microphone measurements. Figure \ref{fig:parallel_levitation} compares the measured and simulated acoustic pressure profiles during parallel execution. The results show consistent pressure distributions across the scanned area, indicating that stable acoustic levitation can be maintained even when two robots operate simultaneously.

\subsubsection{Synchronous Collaborative Manipulation and Acoustic Levitation Stability Analysis (Scenario 3)}

\begin{figure}[!htbp]
    \centering
    \includegraphics[width=0.40\textwidth]{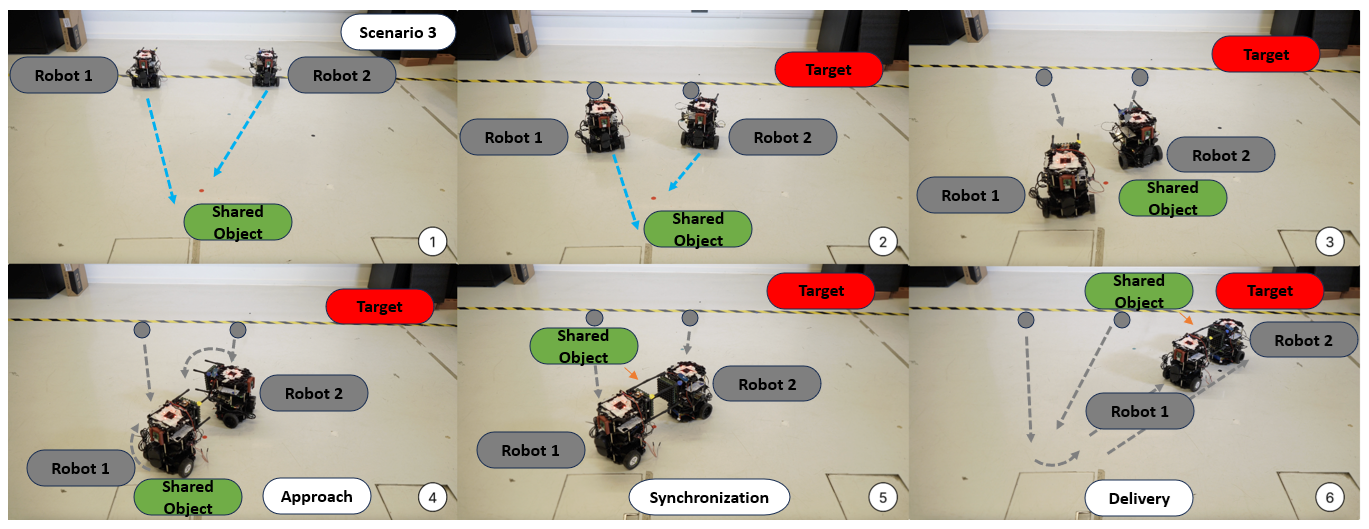}
    \caption{
    Synchronous collaborative task execution (Scenario 3): Two acoustic robots cooperatively transport a shared object using synchronized acoustic manipulation. (1–3) Both robots navigate toward the target location while aligning their relative positions and orientations. (4–6) The robots synchronize their phased-array modules and jointly transport the object toward the designated target position.
    }
    \label{fig:syncTrajectory}
\end{figure}

\begin{figure}[!htbp]
    \centering
    \includegraphics[width=0.40\textwidth]{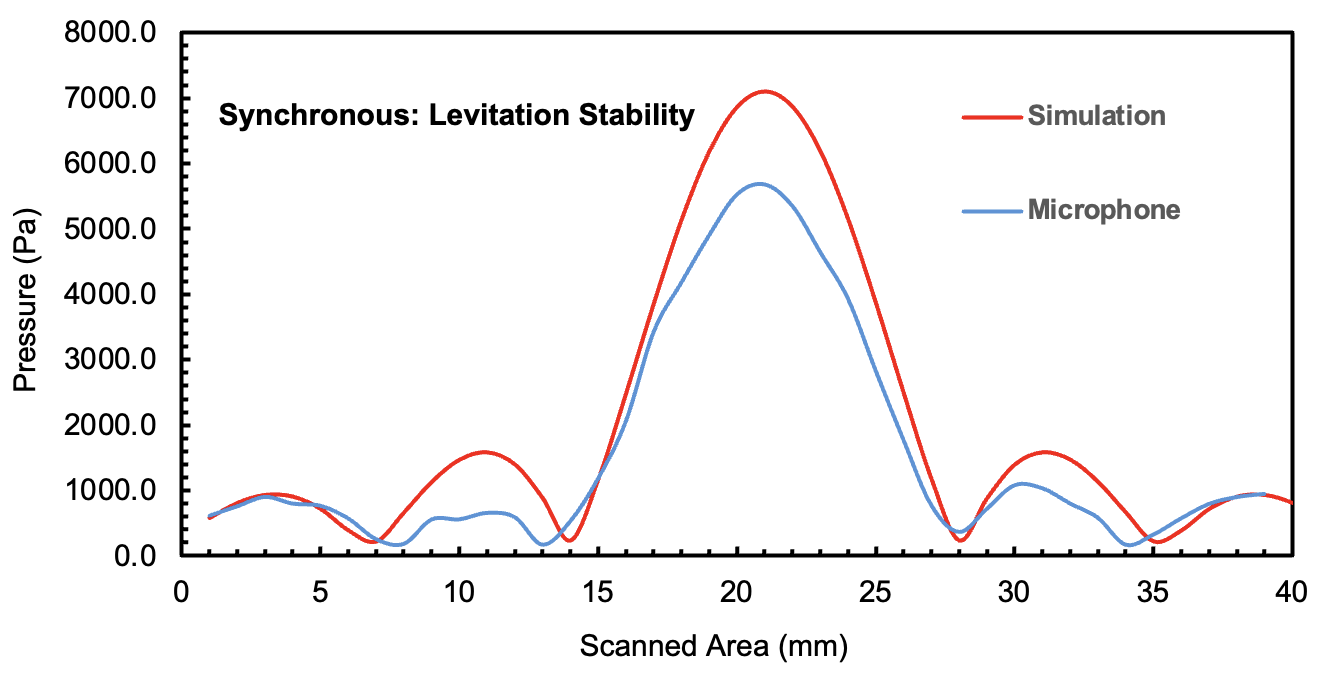}
    \caption{
    Synchronous acoustic levitation stability during cooperative multi-robot manipulation (Scenario 3): The acoustic pressure distribution obtained from MATLAB simulation is compared with microphone measurements during synchronized operation of two acoustic robots, demonstrating stable acoustic field generation for collaborative transport tasks.
    }
    \label{fig:synchronous_levitation}
\end{figure}

The synchronous scenario evaluates the most challenging coordination mode in the framework, where two robots jointly transport a shared object through synchronized acoustic manipulation. Figure \ref{fig:syncTrajectory} illustrates the execution process. Both robots first approach the target region while aligning their relative positions and orientations. After alignment, the robots synchronize their phased-array modules and jointly transport the shared object toward the designated target location.
This experiment demonstrates the feasibility of coordinated acoustic manipulation across multiple robots, but it also highlights the increased difficulty of maintaining stable object transport when both timing and relative spatial alignment must be controlled precisely.
To validate the acoustic field generated during synchronous cooperative transport, MATLAB simulation results were compared with microphone measurements obtained during the experiment. Figure \ref{fig:synchronous_levitation} shows close agreement between the simulated and measured pressure profiles across the scanned region. These results confirm that synchronized phased-array operation can maintain stable acoustic levitation during collaborative multi-robot manipulation.

\subsection{System Performance}

Table \ref{tab:system_performance} summarizes system performance across the three task coordination modes in terms of execution success, parsing accuracy, and latency.
Sequential tasks achieved the highest execution success rate of 96\%, with parsing accuracy of 99\% and latency of 1.2 seconds. Parallel execution achieved 86\% success with 95\% parsing accuracy and 1.8 seconds latency. The synchronous collaborative scenario remained the most challenging, achieving 70\% execution success with 88\% parsing accuracy and 2.5 seconds latency. Failures in the synchronous case were mainly caused by small alignment errors and the sensitivity of the shared acoustic field to relative positioning between robots.


\begin{table}[!htbp]
\centering
\caption{System performance across task execution modes.}
\label{tab:system_performance}

\setlength{\tabcolsep}{4pt}

\begin{tabular}{lccc}
\toprule
\textbf{Task} & \textbf{Execution} & \textbf{Parsing} & \textbf{Latency} \\
\textbf{Mode} & \textbf{Success (\%)} & \textbf{Accuracy (\%)} & \textbf{(seconds)} \\
\midrule
Sequential   & 96 & 99 & 1.2 \\
Parallel     & 86 & 95 & 1.8 \\
Synchronous  & 70 & 88 & 2.5 \\
\bottomrule
\end{tabular}

\end{table}

Overall, these results indicate that the proposed framework maintains high natural language parsing performance across all scenarios, while execution reliability decreases as coordination complexity increases.



\subsection{Natural Language and Speech Interface Performance}

To evaluate the natural language interface, parsing accuracy was measured across different task types. The system maintained high parsing performance across single-robot and multi-robot task descriptions, demonstrating that the LLM-based parser can reliably convert spoken commands into structured task representations suitable for robotic execution.





Speech recognition performance was evaluated using three Whisper model configurations, as summarized in Table \ref{tab:stt_eval}. The Whisper API achieved the highest recognition accuracy (98\%) while also providing lower latency than the locally deployed Whisper Tiny (92\%) and Whisper Base (94\%) variants. The local models offered reduced computational requirements but exhibited slightly lower recognition performance. 

\begin{table}[!htbp]
\centering
\caption{Speech recognition performance comparison across Whisper models used for natural language command interpretation.}
\label{tab:stt_eval}
\begin{tabular}{lcc}
\toprule
\textbf{Model} & \textbf{Accuracy (\%)} & \textbf{Latency (seconds)} \\
\midrule
Whisper Tiny & 92 & 7-9 \\
Whisper Base & 94 & 10-13 \\
Whisper API  & 98 & 5-8 \\
\bottomrule
\end{tabular}
\end{table}

Overall, the combination of Whisper-based speech recognition and LLM-based task parsing provided a reliable interface for natural language-driven multi-robot coordination.

\section{CONCLUSIONS} 
\label{Conclusions}

This paper presented a decentralized multi-agent framework that integrates LLMs with mobile acoustic robots to enable natural language-driven contactless manipulation. The proposed system combines speech recognition, language parsing, distributed task scheduling, and robotic control into a unified architecture that allows non-expert users to interact with multiple robots using intuitive spoken commands.
To support reliable multi-robot coordination, we introduced a structured JSON-based task representation and a distributed scheduling mechanism that can execute sequential, parallel, and synchronous tasks. The system was implemented on TurtleBot3-based acoustic robots equipped with ultrasonic phased arrays for acoustophoretic manipulation of objects. Through the integration of Whisper-based speech recognition, LLM-driven semantic parsing, and natural language commands can be automatically translated into executable robot actions.
Experimental evaluation across three task scenarios demonstrated the feasibility of the proposed framework. Sequential task execution achieved high parsing accuracy and low latency, while parallel multi-robot coordination successfully enabled simultaneous task execution. Synchronous cooperative manipulation highlighted the potential of distributed acoustic transport, although it also revealed challenges related to spatial alignment and timing synchronization between robots.
Overall, the results demonstrate that LLM-based task interpretation can significantly simplify human–robot interaction while enabling flexible coordination in multi-agent robotic systems. The proposed framework shows promise for applications in healthcare logistics, laboratory automation, and contamination-sensitive environments where contactless manipulation is desirable.
This work demonstrates that integrating LLM-driven task reasoning with distributed acoustic manipulation provides a promising foundation for intuitive human–swarm interaction in future autonomous robotic systems.



\section{Acknowledgments}

This work was supported by the EPSRC Prosperity Partnership Program - Swarm Spatial Sound Modulators (EP/V037846/1), and by the Royal Academy of Engineering through their Chairs in Emerging Technology Program (CIET 17/18).

\addtolength{\textheight}{-12cm}   

\bibliographystyle{IEEEtran}
\bibliography{samplebib}


\end{document}